\documentclass{llncs}
\usepackage{makeidx}  
\usepackage{graphicx,times,amsmath,amssymb,bm}
\usepackage[tight,large]{subfigure}
\usepackage{color}
\usepackage[normalem]{ulem}
\begin{document}
\frontmatter          
\pagestyle{headings}  
\addtocmark{Reinforcement learning} 

\mainmatter              
\title{Decision Making Agent Searching for Markov Models in Near-Deterministic World}
\author{G\'abor Matuz and Andr\'as L{\H o}rincz}

\institute{E\"otv\"os Lor\'and University, Budapest H-1117, Hungary,\\
\email{matuzg@gmail.com, andras.lorincz@elte.hu},\\ WWW home page:
\texttt{http://nipg.inf.elte.hu}}

\maketitle              

\begin{abstract}
Reinforcement learning has solid foundations, but becomes inefficient in partially observed (non-Markovian) environments. Thus, a learning agent --~born with a representation and a policy~-- might wish to investigate to what extent the Markov property holds. We propose a learning architecture that utilizes combinatorial policy optimization to overcome non-Markovity and to develop efficient behaviors, which are easy to inherit, tests the Markov property of the behavioral states, and corrects against non-Markovity by running a deterministic factored Finite State Model, which can be learned. We illustrate the properties of architecture in the near deterministic Ms.~Pac-Man game.  We analyze the architecture from the point of view of evolutionary, individual,
and social learning.
\end{abstract}

\section{Introduction}


We are concerned with real world scenarios, where decision making can be arbitrarily poor if available sensory information is insufficient or if some older, yet still ongoing or implicating events are not taken into account. If past information can not improve decision making then state description is Markovian and the problem belongs to the realm of Markov Decision Processes (MDPs), which have solid theoretical foundations. The only remaining question for MDPs is whether the description fits into the memory and whether optimization can be solved in reasonable time or not. The size of the problem grows exponentially with the number of variables. Therefore in order to reduce problem size, state description has to be compressed by extracting relevant features of the decision. As a result, factored reinforcement learning (RL) is obtained, which once again raises the question whether compression spared the Markov property or not.

We are interested in scenarios, when the world is close-to-deterministic. Intriguingly, unless we refer to quantum mechanics, a well-informed Laplace's demon can make decent close-to-deterministic approximations \cite{wolpert08physical}\footnote{Note that we read the theorem backwards.}. Surprisingly, this close-to-deterministic nature may hold --~to a great extent~-- to  everyday human activities, too \cite{song10limits}. With regards to human cooperation, there is a strong correlation between the ability of reading the hidden emotional and cognitive `parameters' of the partners and collaborative (social) skills \cite{frith01mind} indicating the necessity of Markovian state description in behavior optimization. Last, but not least, MDP formalism tacitly includes the concept of determinism: states are perfectly known to the agent within a finite time window. In turn, the concept of state  hides close-to-deterministic finite time processes in MDP.
\begin{figure}[h!]
\centering
    \subfigure[][]{\includegraphics[width=50mm]{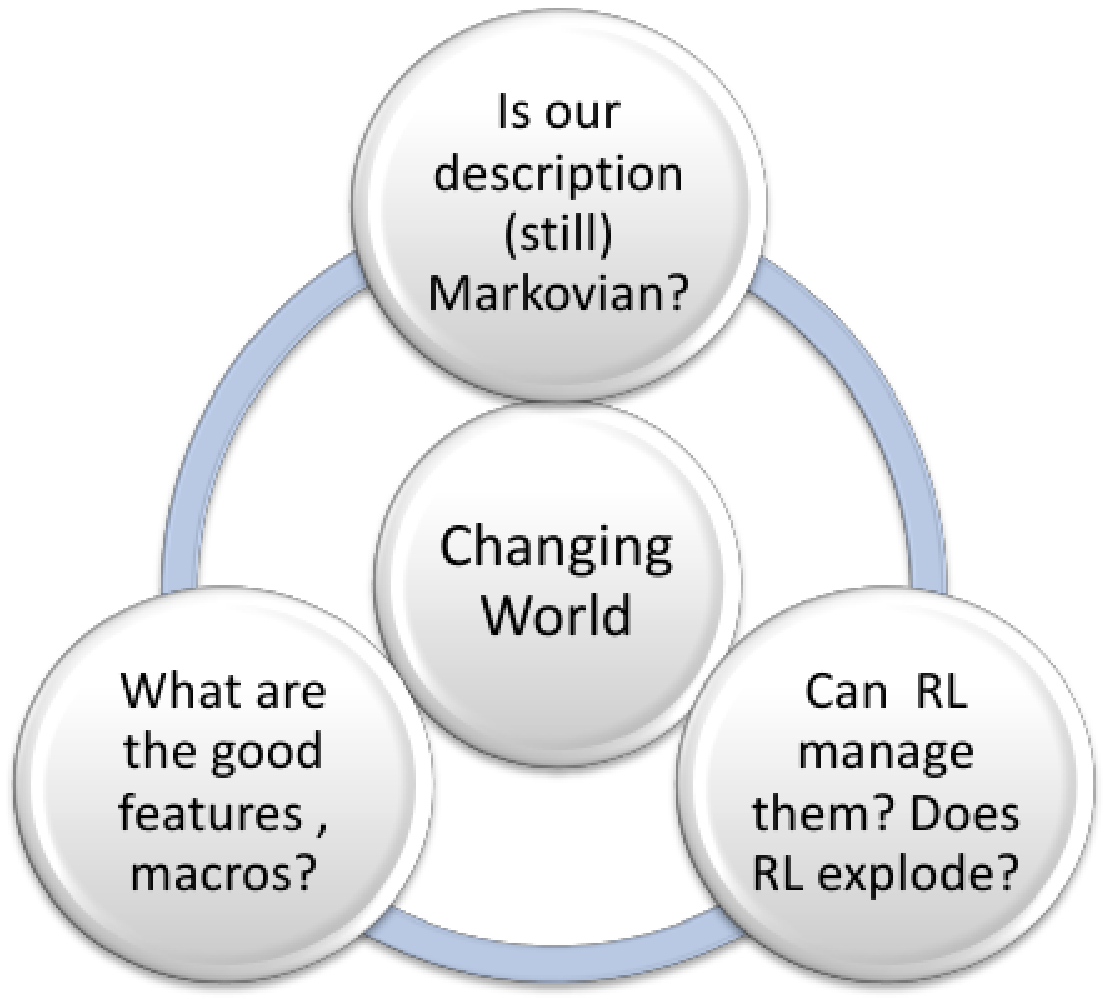}\label{f:learning}}\hfill
    \subfigure[][]{\includegraphics[width=60mm]{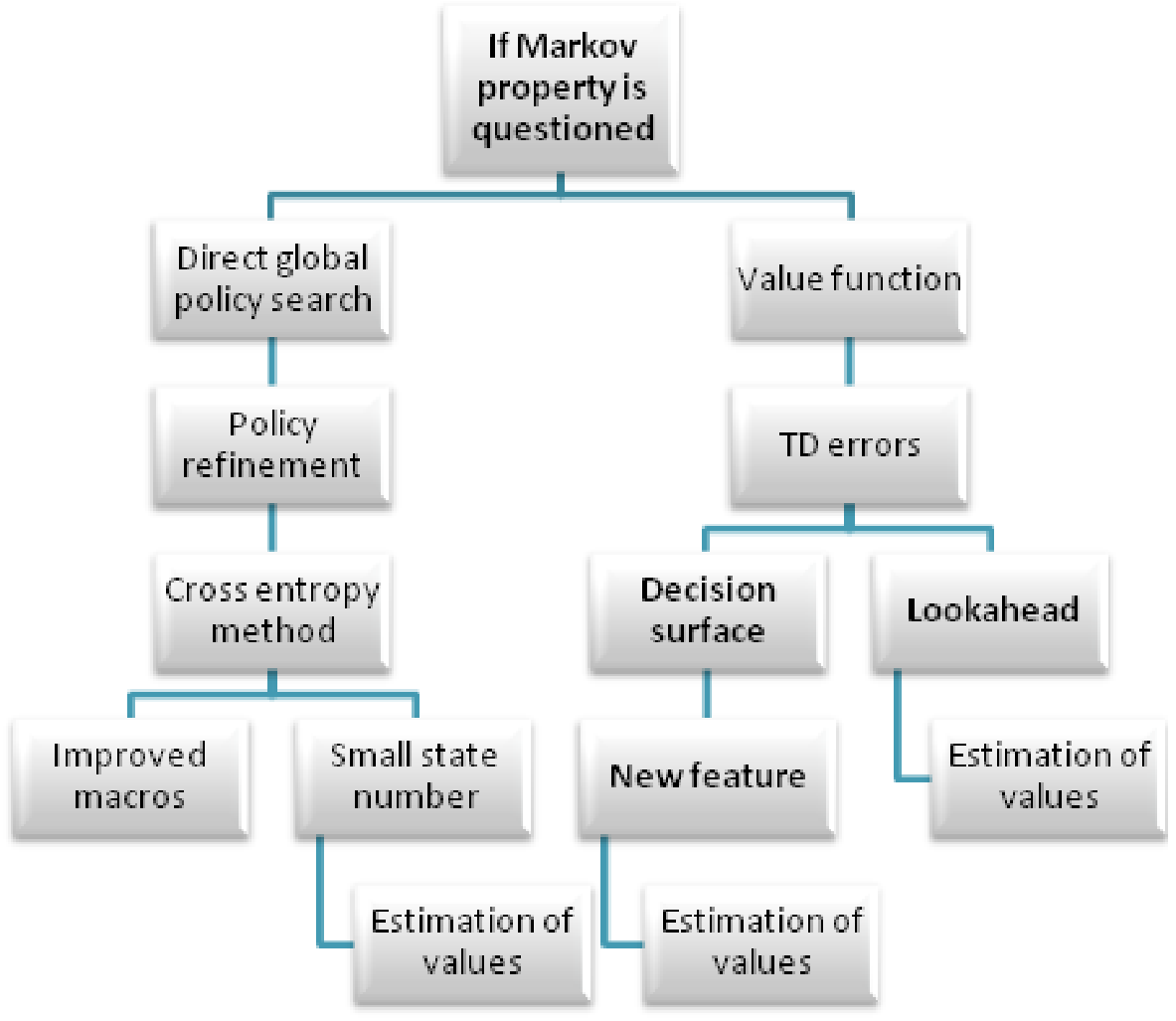}\label{f:DM_non_Markov}}
    \subfigure[][]{\includegraphics[width=120mm]{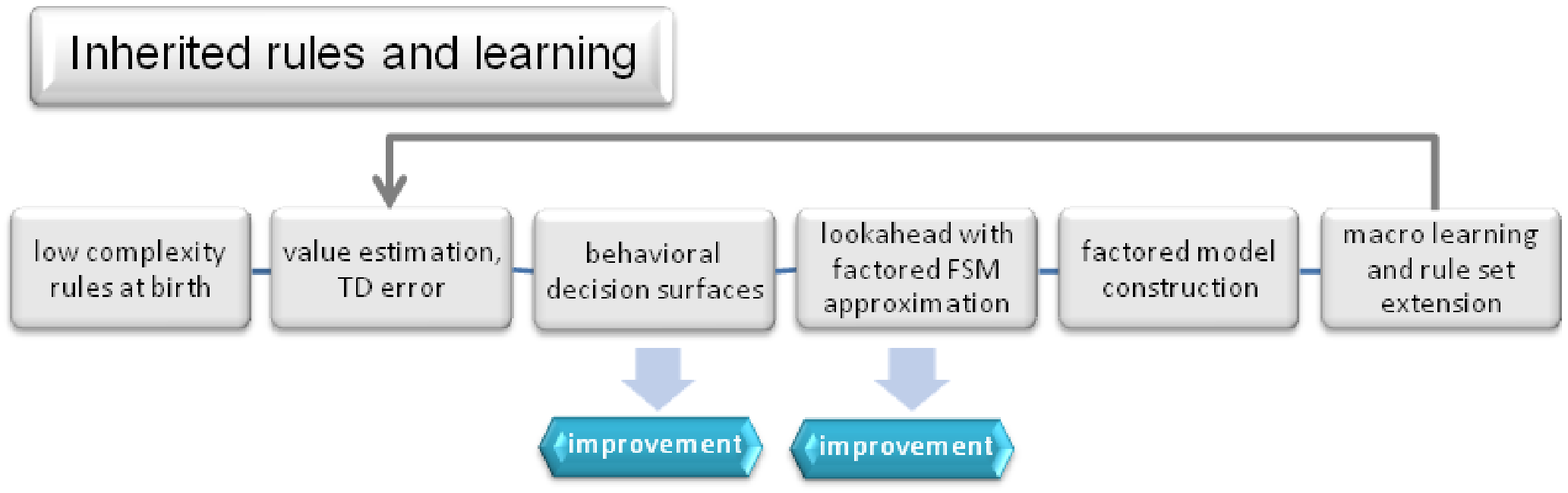}\label{f:architecure}}
\caption{Learning in a continuously changing world. \newline (a): Questions for RL in a changing world, (b): Methods to overcome non-Markovian character of state description, and (c): Architecture for a changing world: agent in is born with a relatively small set of low-complexity rules selected by evolution. Temporal difference (TD) error may uncover non-Markovian character of the behavioral states. Decision surfaces based on TD learning may diminish non-Markovian character. Lookahead method is valuable if factored and close-to-deterministic finite state model of the environment can be learned.}
\end{figure}

Markovian description of the environment is questionable since inherited rules and representations generated by observations may be non-Markovian, or the world may change and the Markovian property may be lost, or agents may change due to the changes of other agents' behavior, and so on.

In turn, we are interested in learning agents. We restrict our considerations to environment, which can be approximated by close-to-deterministic Finite-State Machines (FSMs) or can be factored to such FSMs (fFSM). The practical points are that (i) the so-called factored MDP model  \cite{Boutilier00Stochastic,Guestrin02Efficient} is tractable from the point of view of optimization \cite{szita08factored}, and (ii) near-deterministic processes are relatively easy to learn.

We address the following issues here: Does the problem description possess Markov-property? Is it feasible to solve the underlying learning problem? If answers are negative then which features are to be learned to improve problem description?

These questions are illustrated on Fig.~\ref{f:learning}. Figure~\ref{f:DM_non_Markov} shows certain opportunities
for checking the Markov property without increasing temporal depth. This occurs by considering the difference of experienced and expected approximations of long-term cumulated discounted rewards; the temporal difference (TD) errors. In the Figure, we depicted the optimization for non-Markovian state description by direct global policy search as well as a method that may overcome traps posed by non-Markovian states via look-ahead based value estimation. We will not address the delicate issue of pulling out close-to-deterministic features/factors from high dimensional uncertain observations: this might be the main bottleneck of artificial general intelligence (AGI). Instead, we demonstrate that policy optimization may lead to good performance with non-Markovian behavioral states, look-ahead using close-to-deterministic factors can quickly overcome poor value estimations of those non-Markovian states and can be used to improve decision making.

Figure~\ref{f:architecure} represents our main concepts about a learning agent in a changing world: we assume that the agent is born with a low-complexity rule set that can save her. This rule set can be developed, e.g., by evolution and we use selective global policy optimization. The rule set, however, may lead to non-Markovian states (the behaviors) if the rules are crude or they do not match the actual world. The agent starts to learn by estimating  the values of its behavioral states: it collects information about the \emph{distribution} of TD errors of these states. Using the TD errors, the agent may develop new binary sensors (decision surfaces) in order to improve state description and decision making. Also, the agent may learn close-to-deterministic fFSMs to overcome problems of value estimations for non-Markovian states by utilizing look-ahead methods, since (i) look-ahead is efficient for close-to-deterministic fFSMs and (ii) look-ahead can easily correct poor value estimations.

The paper is built as follows: Markovity and related RL concepts are reviewed in Section~\ref{s:RL}. We executed experimental demonstrations on the Ms.~Pac-Man game in order to illustrate the concepts of  Fig.~\ref{f:architecure}. Ms.~Pac-Man was chosen because of our previous experiences \cite{szita07learning}. In this
paper, considerations are restricted to the single Ms.~Pac-Man case (Section~\ref{s:demo}). However, if proper (mind reading) sensors are available that enable the close-to-deterministic fFSM characterization of the other agent(s), then our methods may have value for the multi-agent case \cite{nipg09turing}.

Results of our experiments are discussed in Section~\ref{s:disc}. We connect our work to theoretically proven favorable features of factored RL, but leave aside the problem of learning the model of the environment. We elaborate on the concept of state in Subsection~\ref{s:state}. We conclude in Section~\ref{s:concout}.


\section{Reinforcement learning}\label{s:RL}

During the last two decades, reinforcement learning has reached a mature state, and has been
laid on solid foundations. We have a large variety of algorithms, including value- function-based,
direct policy search and hybrid methods. For reviews on these subjects, see, e.g.,
\cite{Bertsekas96Neuro-Dynamic,Sutton98Reinforcement,szepes10reinforcement} and references therein. RL deals with sequential decision tasks, where decisions of an agent are not based on
information about locally optimal actions, but its strategy is rather relying on a scalar reward
associated to each state transitions. MDP formalism encapsulates the RL task as follows:

A Markov Decision Process comprise of a quadruple $\{S,A,P,R\}$, where $S$ is a finite set of
states, $A$ is a finite set of actions, $P: S \times A \times S \rightarrow [0,1]$ is a state
transition probability telling the probability of arriving at state $s'$ after executing action $a$
in state $s$. $R: S \times A \times S \rightarrow \mathbb{R}$ is the reward function, $R(s,a,s')$ is the
immediate reward for transition $(s,a,s')$.

The behavior of an agent is described by its policy a $\pi: S \times A \rightarrow [0,1]$ function,
which assigns a probability to an action in each state. For a given policy, each state has an
associated value, which is the expectation value of the long-term cumulated discounted immediate
reward:
\begin{displaymath}
    \mathop{\mathrm{V^{\pi}}}(s) := \mathop{\mathrm{E_{\pi}}}
    \left(
            \displaystyle \sum_{k=0}^{\infty} \gamma^{k} r_{t+k+1}
        \arrowvert
            s_{t} = s
    \right)
\end{displaymath}
where $0 \le \gamma \le 1$ is the discount factor and $r_{t+k+1}$ is the immediate reward in the
$(k+1)^{st}$ step, $\mathop{\mathrm{E_{\pi}}}$ denotes the expectation operator for policy $\pi$.
The goal of an RL agent is to find an optimal policy that chooses the actions giving rise to the
largest values in each state.

When Markov-property holds, the optimal policy ($\pi^*$) is the greedy policy and the corresponding
value function is the optimal value function ($V^*$):

\begin{displaymath}
    \pi^{*}(s) = \mathop{\mathrm{argmax}}_{\displaystyle a}
    \left\{
        \displaystyle  \sum_{s' \in S} P(s,a,s')[R(s,a) + \gamma V^{*}(s')]
    \right\}.
\end{displaymath}

\subsection{Solving reinforcement learning problems}

Many practical applications have appeared that utilize RL. Algorithms with good online performance are known and their asymptotic and finite-state behavior is well understood in most cases.

\subsubsection{Dynamic programming}

When $S$, $A$, $P$, $R$ are available then an iterative method comprised of improvement and
greedification steps can be used to obtain an optimal policy. This approach may fall short when $S$
is large or the model is not available.

\subsubsection{Temporal differences}

Temporal Differences (TD) and its most simple form TD(0) can be used to build an appropriate value
function if matrices $P$ and $R$  are not known or are unmanageably large:

Regarding time instants $t = 0,1,2 ...$ assume that in state $s_{t+1}$ reward $r_{t+1}$ is
observed, while value estimation at time $t$ give rise to $V_t(s_t)$ and $V_t(s_{t+1})$ for states
$s_t$ and $s_{t+1}$, respectively. Then the TD(0) error of the estimation is
\begin{displaymath}
    \delta_{t+1} = r_{t+1} + \gamma V_t(s_{t+1}) - V_t(s_{t})
\end{displaymath}
According to TD(0) learning, value estimation is updated in the following way:
\begin{displaymath}
    V_{t+1}(s) =
        \left\{
            \begin{array}{ll}
                V_{t}(s) + \alpha \delta_{t+1} & \textrm{if} \quad s = s_t \\
                V_{t}(s) & \textrm{if} \quad s \neq \mathop{\mathrm{s_t}}, \\
            \end{array}
        \right.
\end{displaymath}
where $\alpha$ is the learning factor.

Many real problems resist to RL, e.g., if state description is non-Markovian or if the number of variables that
describe the state is large since the size of the state space grows exponentially with the number of variables. Solutions to such problem are considered below.

\subsubsection{Policy search, CE method}


The lack of the Markovian property can be circumvented to some
extent by global policy search methods that do not rely on a value
function but rather an explicit parametric representation of
policies are maintained and updated according to a measure of
rewards acquired during one or more trial episodes. It has been
shown that the theoretically firm Cross-entropy Method (CEM)
\cite{Rubinstein99Cross-Entropy} is efficient in policy search
\cite{Szita06Learning}. Furthermore, CEM can deal with
non-Markovian state-descriptions and the optimization of multiple
actions (action combinations) acting simultaneously, enabling the
optimization of factored action representations
\cite{szita07learning}.

CEM works as follows: Let $ \mathbf{x} = (x_{1},...,x_{n}) \in X$ be a discrete vector and $S: X
\rightarrow \mathbb{R}$ be a so-called goal or fitness function, which defines the fitness of a
particular solution vector $\mathbf{x}$. The goal is to find the vector with the highest
corresponding fitness value:
\begin{displaymath}
    \mathbf{x^{*}} =
        \mathop{\mathrm{argmax}}_{\displaystyle \mathbf{x} \in X}
            \left\{
                S(\mathbf{x})
            \right\}
\end{displaymath}
 CEM maintains a distribution from a family of parametric distributions ($G$) over
the set of possible solutions $X$ and converges towards a distribution in this family whose density
is the highest in an arbitrary environment of the optimal solution $\mathbf{x^{*}}$. The algorithm
has an iterative approach, where $t = 0,1,...,T$ marks subsequent iterations:

In each iteration, let $N$ mark the number of independent sample points
(\mbox{$\mathbf{x}^{(1)},...,\mathbf{x}^{(N)} \in X$}) drawn from the actual distribution $g(t)$.
In case of arbitrary $\theta \in \mathbb{R}$ the set of high-valued sample points is:
\begin{displaymath}
    \hat{L}_{\theta} :=
        \left\{
                \mathbf{x}^{(i)}\arrowvert S(\mathbf{x}^{(i)}) \ge \theta , 1 \le i \le N
        \right\}
\end{displaymath}
which converges to the level set belonging to $\gamma$, i.e., to :
$
    L_{\theta} :=
        \left\{
                \mathbf{x}  \arrowvert S(\mathbf{x}) \ge \theta
        \right\}
$

For high $\theta$ values a uniform distribution over the level set $L_{\theta}$ is concentrated
around the global optimum $\mathbf{x^{*}}$. During the optimization, $\theta$ has to be adjusted
adaptively because at the beginning it is highly likely that the sample set $\hat{L}_{\theta}$ is
empty, for high $\theta$ values. CEM specifies a $\rho \in [0,1]$ ratio, which describes
the number of items in the level set $L_{\theta}$ and therefore determines the actual $\theta$
value. The best $N \times \rho$ items are called elite samples and they typically comprise of
2-10\% of the samples. If the samples are ordered by descending fitness value, then the value of
$\theta$ can be calculated from $\rho$ as $\theta := f(\mathbf{x}^{(\rho N)})$.

After sorting the samples in this way, distribution $g(\mathbf{x})$ can be updated so that the
Kullback-Leibler divergence between a uniform distribution over the approximated level set
$\hat{L}_{\theta}$ and the distribution decreases. For more details, see
\cite{Kroese07Application}.

CEM is a selective batch method and it resembles to genetic algorithms in many respects. CEM can be extended to online optimization \cite{szita08online_TR}, which may suit real world RL problems better.

The exponential blow-up of the state space can be treated with function approximations and factored-state MDPs that are tied together:

\subsubsection{Function approximation}

When the state- or action-space is large or continuous, function approximation methods (FAPPs) can
also be involved, which create value functions by the approximation of sample points while using
supervised learning methods, but now those are controlled by RL. In the case of FAPPs the goal is to create an explicit representation of approximated value functions and parameters are typically updated after each consecutive step.
For convergent general algorithms see \cite{icml2010_092} and the references therein. Special FAPPs
that have temporal dynamics can also be used to diminish consequences of non-Markovian observations
\cite{Szita06Reinforcement}.

\subsubsection{Factored-state Markov Decision Processes}

We assume that $S$ is the Cartesian product of $m$ smaller state spaces (corresponding to
individual variables):
\[
  S = S_1 \times S_2 \times \ldots \times S_m,
\]
where $S_i$ has the size $|S_i| \ll |S|$ ($i=1, \ldots , m$). This is the case of factored Markov
Decision Process (fMDP, or factored RL)
\cite{Koller00Policy,Boutilier95Exploiting,Boutilier00Stochastic}. A naive, tabular representation
of the transition probabilities would require exponentially large space (that is, exponential in
the number of variables $m$). However, the next-step value of a state variable often depends only
on a few other variables, so the full transition probability can be obtained as the product of
several simpler factors. It has been shown that by uniformly sampling polynomially many samples
from the (exponentially large) state space, the complexity of the fMDP solver stays polynomial in
the size of the fMDP description length, i.e. in $\sum_{i=1}^m |S_i|$ \cite{szita08factored}. This
convergent fMDP solver makes use of function approximations and ties fMDPs to FAPPs.

Factored MDPs may also take advantage of the module concept of RL, see below.

\subsubsection{RL modules}

The concept of time step is very flexible in MDPs. A particular solution defines time intervals by
actions: the state is described by the available modules, i.e., the complex actions that can be
used at a given time. Then state changes only when the set of available modules changes and the new
set defines the new state, whereas the activated module of the set defines the action in that state
\cite{Kalmar98Module-Based}. This formulation is advantageous for two reasons. For one, the formulation
can easily be extended by robust controllers \cite{Szita03Epsilon-MDPs}. In addition the formulation enables to activate more than one module --~a macro~-- at a time \cite{szita07learning} giving rise to combinatorial flexibility, factored descriptions, and behavioral states upon optimization as we shall see it later.

\subsubsection{Model construction and planning in RL}

Learning of the transition probability matrix, the dynamical model of RL, offers several advantages. First of
all, this model enables planning. It is useful if value estimation is imprecise, for example if rewards may disappear, or new rewards may appear. In this case, one can run the learned model, like in dynamic
programming. Typical usage of the model computes the expected value of a state by summing up discounted future rewards according to the dynamical model and the policy. Advantages appear if real world experimentation is limited, and some of the rewards, certain transition probabilities, or a few of the factors  may change, but can be learned easily. Upon observation of such changes, the dynamical model can update value estimation without further real world experiments.

The other advantage of model learning is that the exploration exploitation dilemma can be resolved
under certain conditions, both for RL \cite{szita08many} and for factored RL \cite{szita09optimistic-short}.

\subsubsection{Factored Finite State Model approximation for planning}

Planning in the Ms.~Pac-Man game (see later \cite{szita07learning}) is severely restricted by the size of the
state space. On the other hand, the state transition can be easily described with factors. For example, if Ms.~Pac-Man
walks over a dot then it disappears. Learning of this `machine' is very simple. Walls and corridors of the game are also simple `machines' since they do not change. There are other, somewhat more complex machines in the game; they have a bit of probabilistic behavior. All of these components are simple to learn and thus an approximate sampling based \cite{kearns99sparse} or a risk avoiding \cite{banerjee07general} (close-to-)deterministic (factored) model can be run by Ms.~Pac-Man for a low cost and with low memory demand for the sake of look-ahead (planning).

Note that look-ahead is typical in games, including Ms.~Pac-Man (see e.g. \cite{thompson08evaluation}) and it is also considered in the MDP literature (see, e.g., \cite{givan03equivalence} and the references therein). Here, we consider look-ahead from the point of view of a general RL architecture that inherits knowledge, learns, and faces new challenges. There are many methods to solve the FSM identification task \cite{Bouloutas93Fault}, provided that the close-to-deterministic features are available.

\section{Architecture through an illustrative example}\label{s:demo}

In this section we illustrate the logic of the algorithmic components as shown in Fig.~\ref{f:architecure}. We assume
a reinforcement learning scenario, where an MDP possesses a large action- and/or state-space but no
method is known to efficiently compress the information via feature extraction, while preserving
the Markov-property of the description. We use the Ms.~Pac.Man game for the illustration.

\subsection{Pac-Man and reinforcement learning}
\label{s:pacman_and_rl}


\begin{figure}
\centering
\includegraphics[width=5cm]{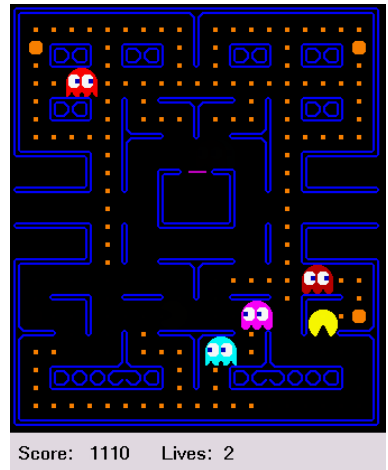}
 \caption{A snapshot of the Pac-Man game}\label{fig:pacman_board}
\end{figure}

The video-game Pac-Man was first released in 1979, and it is considered to be one of the most popular
video games to date \cite{Wiki:Pac-Man}.

The player maneuvers Pac-Man in a maze (see Fig.~\ref{fig:pacman_board}), while Pac-Man \emph{eats}  the dots in the maze. In this particular maze there are 174 dots, each one is worth 10 points. A level is finished when all the dots are eaten. There are also four \emph{ghosts} in the maze who try to catch Pac-Man, and if they succeed, Pac-Man loses a life. Initially, he has three lives, and gets an extra life after reaching 10,000 points.

There are four power-up items in the corners of the maze, called \emph{power dots} (worth 40 points). After Pac-Man eats a power dot, the ghosts turn blue for a short period (15 seconds), they slow down and try to escape from Pac-Man.
During this time, Pac-Man is able to eat them, which is worth 200, 400, 800 and 1600 points, consecutively. The point values are reset to 200 each time another power dot is eaten, so the player would want to eat all four ghosts per power dot. If a ghost is eaten, his remains hurry back to the center of the maze where the ghost is reborn. At certain intervals, a fruit appears near the center of the maze and remains there for a while. Eating this fruit is worth 100 points.

We restricted our studies to the first level, so the maximum achievable score is $174\cdot10 + 4\cdot40 + 4\cdot(200+400+800+1600) = 13,900$ plus 100 points for each time a fruit is eaten.

In the original version of Pac-Man, ghosts move on a complex but deterministic route, so it is possible to learn a deterministic action sequence that does not require any observations. In \emph{Ms.~Pac-Man}, randomness
was added to the movement of the ghosts. This way, there is no single optimal action sequence, observations are necessary for
optimal decision making. We used the implementation available on the Internet \cite{PacMan_CIG}.

\subsubsection{Ms.~Pac-Man as an RL task}

Ms.~Pac-Man meets all the criteria of a reinforcement learning task. The agent
has to make a sequence of decisions that depend on its observations. The
environment is stochastic (because the paths of ghosts are unpredictable).
There is also a well-defined reward function (the score for eating things), and
actions influence the rewards to be collected in the future.

The full description of the state would include (1) whether the dots have been
eaten (one bit for each dot and one for each power dot), (2) the position and
direction of Ms.~Pac-Man, (3) the position and direction of the four ghosts,
(4) whether the ghosts are blue (one bit for each ghost), and if so, for how long they remain blue (in the range of 1 to 15 seconds) (5) whether the fruit is present, and the time left until it appears/disappears (6) the number of lives left. The size of the resulting state space is huge, so some kind of function approximation or feature-extraction is necessary for RL.

The action space is much smaller, as there are only four basic actions: go
north/south/east/west. However, a typical game consists of multiple hundreds of
steps, so the number of possible combinations is still enormous. This indicates
the need for temporally extended actions.

We provide mid-level domain knowledge to the algorithm: we use domain knowledge to preprocess the state information and
to define action modules. On the other hand, it will be the role of the policy search reinforcement learning to combine the observations and modules into rule-based policies and find their proper combination.

Ms.~Pac-Man is attractive for the illustration, because the environment --~alike to many other games~-- has some life-like characteristics, and the agent has to solve interrelated tasks and identify interconnected behaviors in a changing environment. These interrelated tasks are not separable and have mutual effects on each other.

The purpose of our demonstration is to assemble a common sense RL architecture by utilizing certain RL techniques. We shall argue later that feature extraction is the main bottleneck for this architecture and shall consider what kind of features are advantageous for RL.

\subsection{Development of rule-based policies and learning macros}

A rule-based policy is a set of rules with some mechanism for breaking ties, i.e., to decide which
rule is executed, if there are multiple rules with satisfied conditions. Actions can last for a
while and some of them may be applied in combinations. We reproduced the actions of
\cite{szita07learning}, examples being like
\begin{itemize}
  \item[(i)] \texttt{ToDot}: Go towards the nearest dot,
  \item[(ii)] \texttt{ToPowerDot}: Go towards the nearest power dot,
  \item[(iii)] \texttt{FromPowerDot}: Go in direction opposite to the nearest power dot,
  \item[(iv)] \texttt{ToEdGhost}: Go towards the nearest edible ghost,
  \item[(v)] \texttt{FromGhost}: Go in direction opposite to the nearest ghost,
\end{itemize}
among others. Here are some examples for observations used for rule constructions. For the full list, see the original publication \cite{szita07learning}:
\begin{itemize}
  \item[(i)] \texttt{NearestDot}: Distance of nearest dot,
  \item[(ii)] \texttt{NearestPowerDot}: Distance of nearest power dot,
  \item[(iii)] \texttt{NearestGhost}: Distance of nearest ghost,
  \item[(iv)] \texttt{NearestEdGhost}: Distance of nearest edible (blue) ghost.
\end{itemize}

Observations can be extended to conditions. For the sake of simplicity, conditions were restricted to have the
form  {\texttt{[observation] < [value]},} {\texttt{[observation] > [value]},} {\texttt{[action]+},} {\texttt{[action]-},}
or the conjunction of such terms. For example, $$\mbox{\texttt{(NearestDot<5) and (NearestGhost>8) and (FromGhost+)}}$$ is a valid condition for our rules.

Once we have conditions and actions, rules can be constructed easily. In our implementation, a rule
has the form ``\texttt{if [Condition], then [Action]}.'' For example, $$\mbox{\texttt{if (NearestDot<5) and
(NearestGhost<5)}\texttt{ then FromGhost+}}$$ makes a potential rule. Each rule has a priority assigned. When the agent has to make a decision, she checks her rule list
starting with the ones with highest priority. If the conditions of a rule are fulfilled, then the
corresponding action is executed, and the decision-making process halts.

\begin{figure}
\centering
    \includegraphics[width=10cm]{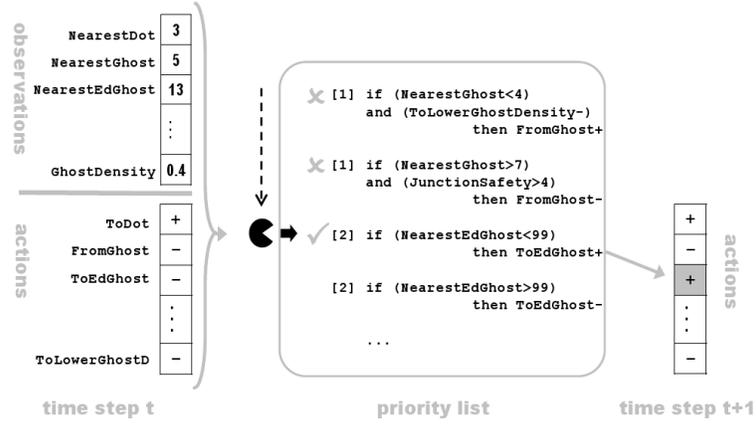}
\caption{Decision-making for Ms.~Pac-Man agent. At time step $t$, the she receives the actual
observations and the state of her action modules. She checks the rules of her priority list in
order, and executes the first rule with satisfied conditions. Plus sign means that an action module is enabled.}\label{f:pacmanrules}
\end{figure}

If a rule with priority $k$ \emph{switches on} an action module, then the priority of the action
module is also taken as $k$. Intuitively, this makes sense: if an important rule is activated, then
its effect should also be important. If a rule with priority $k$ \emph{switches off} a module, then
it is executed, regardless of the priority of the module. The procedure is depicted in Fig.~\ref{f:pacmanrules}.
Rules can be pre-wired, or can be created randomly, and a global search method can select the good
ones and assign their priorities \cite{szita07learning}. Thus, learning can be concerned with (i)
combining different conditions and actions to form rules and (ii) deciding about the priorities of
the rules. This policy optimization is accomplished by the cross-entropy method. \emph{Macros} are simultaneously used rule combinations.

The optimization procedure gives rise to a `low-complexity policy': there can be many rules in the
policy but only a few rules can be concurrently active at a time. The effective length of policies
is biased towards short policies. Frequent, low-complexity rule combinations can be used as
building blocks in a search for a more powerful but still low-complexity policies. The result, i.e.
the optimized Ms.~Pac-Man policy may be seen as a new action that can be selected under certain
conditions. For more details on the rules, see \cite{szita07learning}.

\subsection{Score, macro values and macro sequences}

\subsubsection{Score} Global policy optimization gave rise to good performance: we used only the first level and the average score was 6680 and the average life was 2.82, out of the 3 possible lives. The maximum achievable score is $174 \cdot 10 + 4 \cdot 40 + 4 \cdot (200 + 400 + 800 + 1600) = 13,900$. Ms.~Pac-Man was not able to eat all ghosts after eating a power dot. From the score it follows that she could eat about 2 to 3 ghosts after consuming the power dot.

\subsubsection{Behavioral states} Macros correspond to a group of simultaneously active rules and can be interpreted as behavioral states or behavioral patterns with the action equivalent to the application of the actions of the macro. A very little set of macros emerged in the Ms.~Pac-Man game, there were only 6 macros. They are listed in Table~\ref{t:macro_values}. The values of the behavioral states can be estimated, e.g., by the TD method.
\begin{table}[h!]
    \centering
    \begin{tabular}{ c c c p{48mm} }
        \hline\noalign{\smallskip}
         \,\, No. \,\,\,\,  Code\,\,\,\,\,\, & \,\, Score \,\, & $\,\,\,\,V(s)\,\,\,\,$  & $\,\,\,$Description \\
        \noalign{\smallskip}\hline\noalign{\smallskip}
            1. \,\,\, 010011 & 5133 & 6621 & \mbox{\texttt{ FromGhost- \& FromPowDot}}\\
            2. \,\,\, 010101 & \textbf{\emph{6812}} & 7024 & \mbox{\texttt{ FromGhost- \& ToEdGhosts}} \\
            3. \,\,\, 011001 & 4006  & 6606 & \mbox{\texttt{ FromGhost- \& ToPowDot}} \\
            4. \,\,\, 100011 & \textbf{\emph{6732}}  & 6142 & \mbox{\texttt{ FromGhost+   \& FromPowDot}} \\
            5. \,\,\, 100101 & 6024 & 6455 & \mbox{\texttt{ FromGhost+  \& ToEdGhosts}} \\
            6. \,\,\, 101001 & 1770 & 6100 & \mbox{\texttt{ FromGhost+  \& ToPowDot}} \\
        \noalign{\smallskip}\hline\smallskip
    \end{tabular}
    \caption{Behavioral states, or macros, their codes and their values. \newline \mbox{$1^{st}$ \emph{column:}} number of behavioral states. $2^{nd}$ \emph{column:} individual digits denote if an action is `on' or  `off'.  Meaning of the digits from left to right: \texttt{FromGhost+}, \texttt{FromGhost-}, \texttt{ToPowDot}, \texttt{ToEdGhosts}, \texttt{FromPowDot}, and \texttt{if Constant>0 then ToNearestPill+}. The last digit of the code applies under all conditions.  $3^{rd}$ \emph{column:} Ms.~Pac-Man scores after CEM optimization of actions \emph{during the time} when behavioral state is effective. Averages for 5000 runs. \textbf{\emph{Bold italic}}: optimization of the actions of the macro increased the score above 6680 (the average score of inherited behavior) and policy greedification is possible. $4^{th}$ \emph{column:} estimated values of inherited behavioral states $V(s)$ upon optimization of the actions. Averages for 5000 runs. \mbox{$5^{th}$ \emph{column:}} description of the behavioral states.}\label{t:macro_values}
    \vskip -6pt
\end{table}

Since Table~\ref{t:macro_values} is very small, it is easy to pre-wire these rules. Since CEM is a probability based, selective, global optimization method, we consider it as the model of evolution here.  These pre-wired rules and their respective priorities give good chances for the agent to survive for a while and to collect additional information to improve performance via individual learning. We note that thought experiments may also apply CEM, but CEM may be too dangerous for experimentation.

\begin{figure}[t!]
\centering
    \subfigure[][]{\includegraphics[width=120mm]{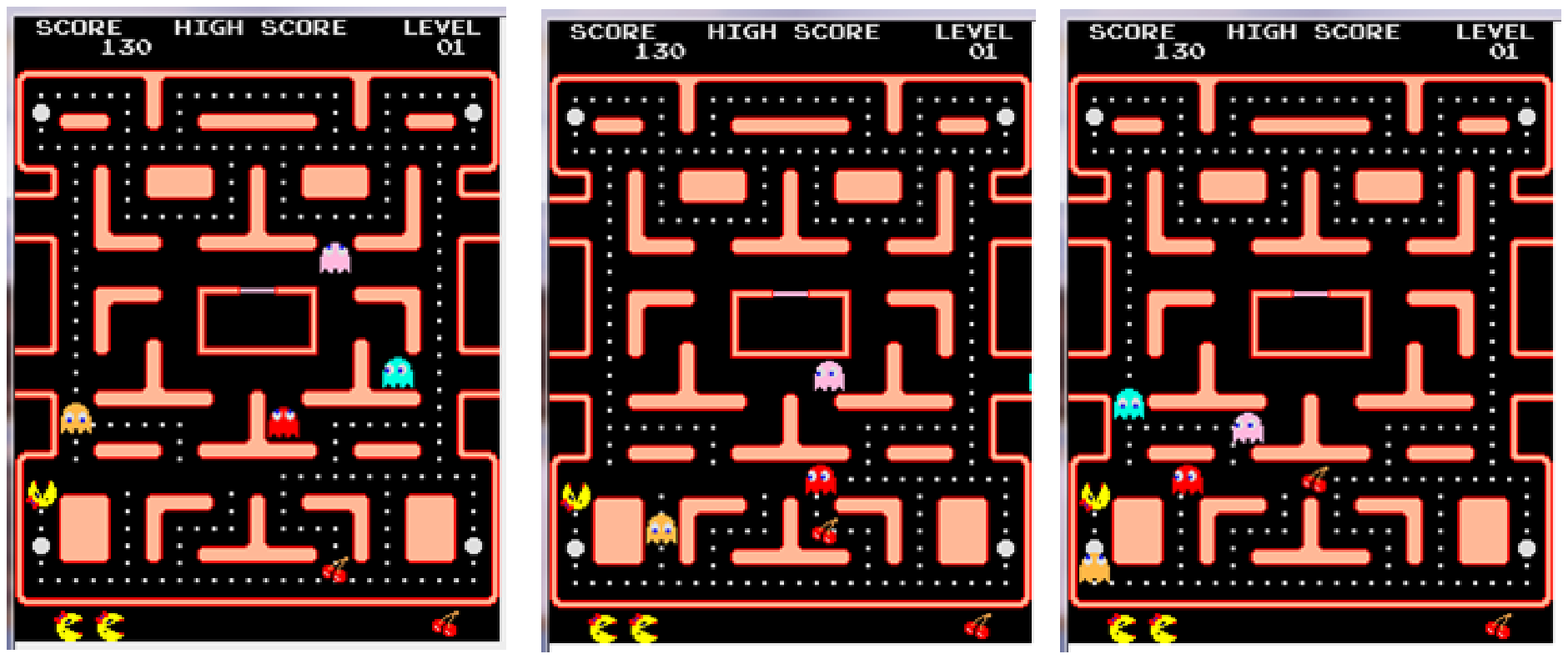}\label{f:Trap1}}
    \subfigure[][]{\includegraphics[width=120mm]{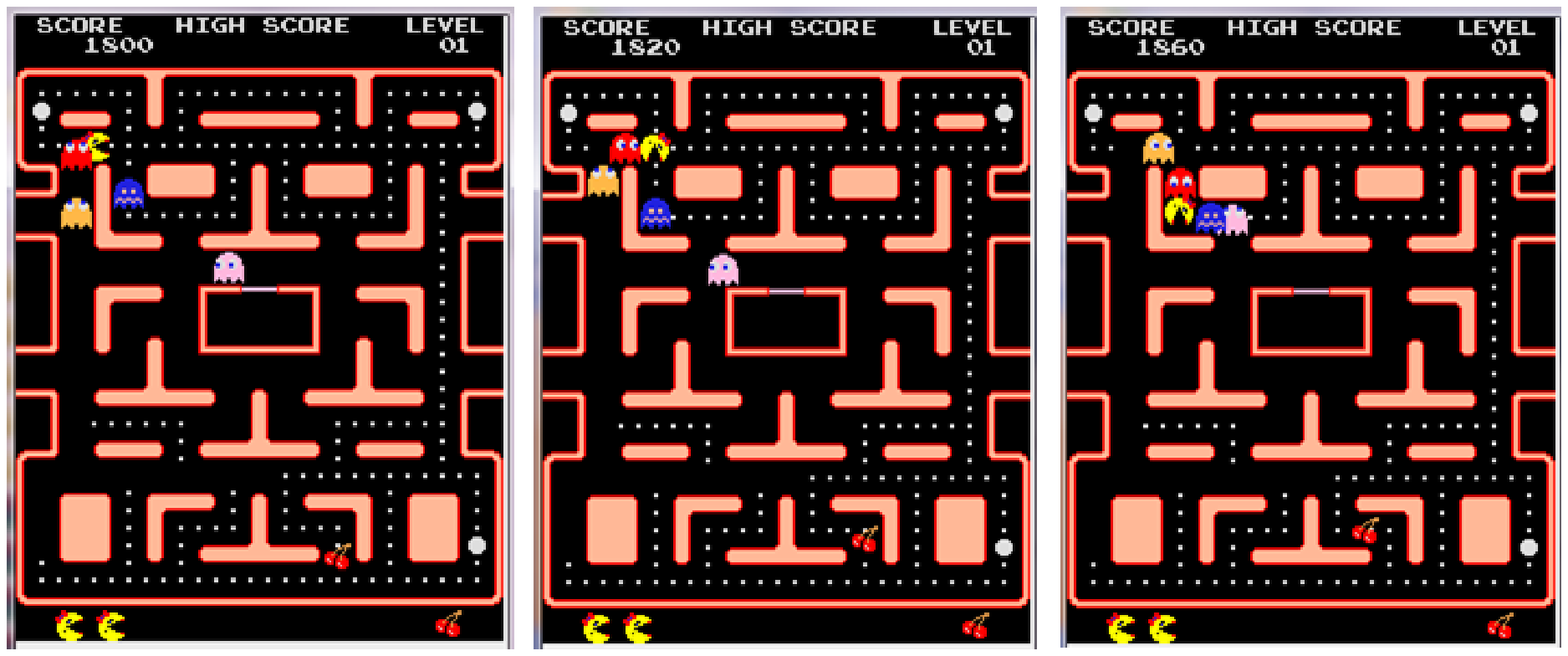}\label{f:Trap2}}
\caption{Lack of look-ahead. \newline (a) Clyde chooses a random move and moves to the right instead of chasing Ms.~Pac-Man [left] and [middle]. Then Clyde switches behavior, approaches Ms.~Pac.Man, and comes form the troubled direction: it crosses the power dot and then she can catch Ms.~Pac-Man [right] \newline (b) [left]: Ms.~Pac-Man could escape, [middle]: She chases the edible ghost, and [right]: falls into the trap closed by Pinky.   }\label{f:pacmantrap}
\end{figure}

Some macros often occur consecutively and could be joined into macro sequences. The formation of macro sequences means a higher level of abstraction, which basically incarnates a new behavior. An example of this phenomenon is the observed zig-zagging
behavioral pattern, when the Ms.~Pac-Man agent performs turns towards and from the power dot in each step: out of the many potential choices, CEM selected a rule set at the second priority level:
\begin{itemize}
  \item[P2:] \texttt{if GhostDensity<1.5 and NearestPowerDot<5 then FromPowerDot+}
  \item[P2:] \texttt{if NearestEdGhost>99 then ToPowerDot+}
\end{itemize}
That is, if ghost density is low and the power pill is close then move away from the power pill. If this rule is on, then the other rule is considered: if there is no edible ghost then move towards the power pill. These two rules at priority level 2 achieve good performance since they give rise to a zig-zagging behavior close to the power pill until ghost density is small. It looks like as if the Ms.~Pac-Man were waiting for the Ghosts and to eat the power dot when they approach in order to have a better chance to catch them afterwards. This is an emerging behavior, since there is no rule for `waiting'. It is unrelated to foreseeing that ghosts may come closer in the future, it is simply the result of selection under global policy optimization. Zig-zagging may also gives rise to a trap as shown in Fig.~\ref{f:Trap1}.

There are proper conditions for the zig-zagging behavior, so one could turn zig-zagging ON and OFF and zig-zagging could be a new macro. The repetitive behavioral pattern can be easily identified during execution and it is easy to compress it into a new module that may undergo CEM optimization (but see Lamarckism). In the following, we consider individual learning during a small number of episodes, i.e., during the game.

\subsection{Individual Learning}

Policies and macros can be optimized further e.g. by:
\begin{enumerate}
    \item considering each macro state as a sub-problem as long as the macro is `ON' and applying CEM during this time interval,
    \item computing the values of the states, inspecting the TD error histogram and then building up predictors
    and decision surfaces for large TD errors and thus increasing the `state space'. Then, using the new states and using the rule set, a new CEM optimization may improve performance.
    \item Alternatively, we may apply a look-ahead procedure.
\end{enumerate}
Note that the first option develops a new behavior for each state without using additional information. This method can help since it extends the range of priorities by introducing new priority set within each `state'. The other two options take advantage of additional information collected during the game. We consider the three options below.

\subsubsection{Subtask optimization}

The optimization of individual macro states via CEM may give rise to novel behaviors that collect more rewards in each state. We executed this program and could increase the collected rewards in each state of the Ms.~Pac-Man game. Then --~if Markov condition is satisfied~-- greedy replacement of the new behavior should improve performance. However, there is no warranty for improvement if the Markov condition is not satisfied. The third column of Table~\ref{t:macro_values} shows average performance after greedy behavior optimization in each state. Since average performance equals to 6680 points, performance improves only by changing behaviors in state 2 and state 4 and not in the other states. For example, optimizing behavior for state 6, we get very poor performance (1770 points on average) upon greedy policy optimization.

Subtask optimization does not suit individual learning well, since it applies CEM and is thus slow. It is also dangerous since overall performance may easily drop even if optimization of behavior is successful.

\subsubsection{Decision surface using TD errors}

\begin{figure}[h!]
\centering
    \subfigure[][Histogram of TD errors]{\includegraphics[width=90mm]{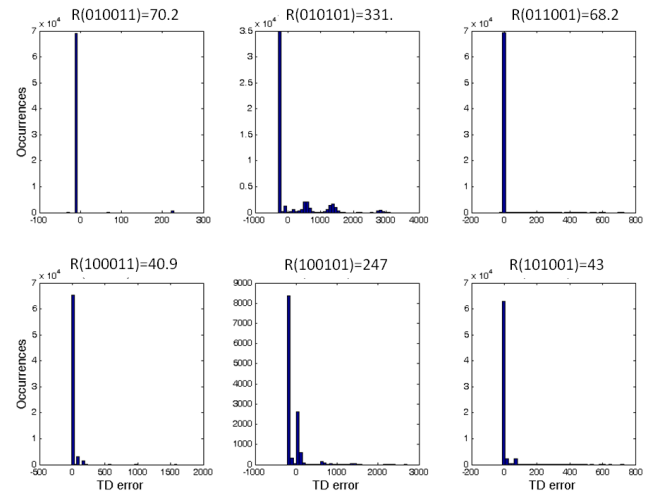}\label{f:TDfull}}
    \subfigure[][Histogram of TD errors \emph{without} the main peak]{\includegraphics[width=90mm]{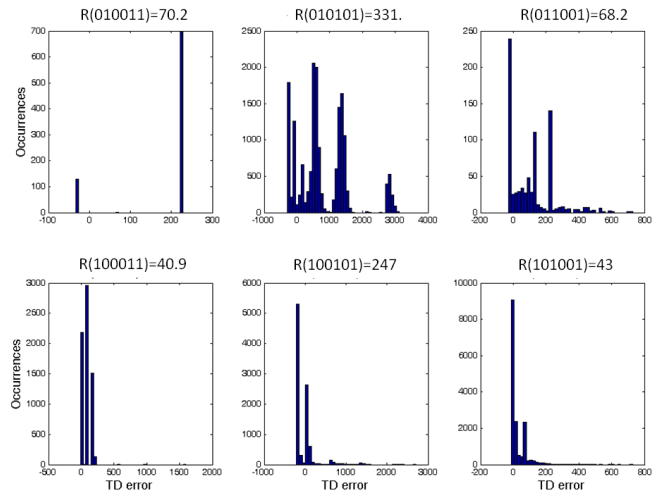}\label{f:TDPart}}
\caption{TD error distribution. Distributions are uneven indicating that decision making can be improved.}\label{fig:TDerror}
\end{figure}

We have estimated the values of the behavioral states (Table~\ref{t:macro_values}, fourth column) and computed the histogram of TD errors (Fig.~\ref{fig:TDerror}). Large TD errors occur (Fig.~\ref{f:TDfull}) that we made visible by excluding the most frequent TD error that determine behavioral success (Fig.~\ref{f:TDPart}). Consider state 2, for example. The average reward collected in this state is about 330. This is cumulated from dots and ghosts. The largest TD peak is at around -150, i.e., in this case Ms.~Pac-Man collected about a single ghost and very few dots. The second peak is at around 600 and it is broader. This corresponds to eating two ghosts and about 30 dots. The third and the fourth peaks are at around 1,400 and 2,800 meaning that Ms.~Pac-Man could collect three or four ghosts, respectively.

It would be ideal for Ms.~Pac-man to avoid TD errors -150, 600, and 1,400 and to collect all four ghosts. One may build a classifier for this purpose by collecting situations when all four ghosts will be captured and all other situations. The classifier then forecasts the reward and defines an RL problem, which either wins (reaches the goal) or looses. CEM can be applied to the optimization of actions starting from a number of situations collected during the game. This procedure is a special form of goal-related feature extraction. It can be solved by RL methods and we did not execute it. The situation corresponds to categorical perception (see, \cite{harnad87category} and references therein), which is to be favored if factored description is not available. We will come back to this point later. We will argue that searching for close-to-deterministic factors, applying factored FSM approximation together with look-ahead is more flexible.

\subsection{Look-ahead module}

In the previous section, we argued that decision surfaces should help decision making. However, there are potential problems with decision surfaces even in the Ms.~Pac-Man game that we list here.
\begin{description}
  \item[World may change:] Considerable part (if not all) of the knowledge that we encode into the decision surface may be lost if the world changes. There are many variables of the world and changes to any of them may spoil decisions. We note that human recognition utilizes both component based recognition, when decision about a category involves the presence or absence of the components \cite{biederman87recognition}, and categorical perception, or example based learning, when sensory observation is changed around the decision surfaces (see e.g., \cite{harnad87category}). This latter version of learning appears only if component based recognition is not useful. Component based recognition seems easier. As an example, consider face recognition. The hierarchical structure of the components, i.e., the components of the face do not change when the environment changes, e.g., if light condition changes. Furthermore,  component based recognition works also under occlusion.
  \item[Learning is hard:] It is hard to learn to learn decision surfaces in real life scenarios. Even in the artificial example of Ms.~Pac-Man a slight change of the corridors of the labyrinth or the actual number, position, and type of the Ghosts may make a big difference. Such sensitivity to small changes of individual variables or combinations of the variables may be be compensated by an enormous number of training examples, but it becomes prohibitively expensive. Whereas the component based recognition suits high dimensional problems and it is typically easy to learn, categorical perception is hard to learn. Categorical perception appears mostly in problems, where the inherent dimension is low. Examples include decisions about female or male faces, borders between colors, and so on.
\end{description}
Taken together, example based learning of decision surfaces in the Ms.~Pac-Man problem is fragile and takes long.

It is however, easy to learn the behavior of the components. For example, any brick of the wall is a component in the Ms.~Pac-Man game. Bricks have coordinates and they do not move. Dots are different: they also have coordinates, but they disappear when Ms.~Pac-Man crosses them. Power dots are similar, but they also change the behavior of the ghosts. Ghosts are either escaping or they are chasing Ms.~Pac-Man. State transitions are simple, but chasing behavior has some randomness. Fruits (that we did not use) appear randomly, they have close-to-deterministic trajectories: upon observation, a decision could be made if the fruit trajectory can be intercepted or not.

Thus, sophistication of the Ms.~Pac-Man game comes from the configurational diversity, whereas the behavior of the components are very simple, sometimes deterministic, and they are easy to learn "on the fly". Learning of the dynamics of the components enables factored description and produces a factored world model with low branching ration that can produce potential configurations. Changes in the world, like different labyrinth, or different ghost speed are again easy to observe and learn. Finally, since branching ratio is low, such models are easy to run and can be used for look-ahead (LA). LA is commonly used in RL, including games (see, e.g., \cite{kearns99sparse,banerjee07general,thompson08evaluation}).

Here, we implemented and non-perfect fFSM model, a model of the game.\footnote{We have bypassed the learning of the fFSM, which is very simple for Ms.~Pac-Man and would be misleading. We believe that pulling out the close-to-deterministic features is the most relevant piece of the learning problem. This problem is treated elsewhere \cite{lorincz11Markov}} Look-ahead used this game model and was implemented as a set of trajectories, where each trajectory contained a future state and the rewards collected up to that future state. Trajectories were enumerated using a depth-first search strategy applied on the state graph of the original model. If look-ahead model approximates the world closely, then it can improve policy starting at (any) state $s_0$:
\begin{eqnarray}
    \pi^{LA}(s_0) & = & \mathop{\mathrm{argmax}}_{\mathrm{LA\,\, trajectory}} \,\,\,\,\, \sum_{\mathrm{LA\,\, trajectory}}P(s_0,a_0,s_1) \, [R(s_1,a_1  \nonumber \\
    & &  + \, P(s_1,a_1,s_2)  \,  [R(s_2,a_2 + \ldots    \nonumber \\
    & & + \, P(s_{T-1},a_T,s_T) \, [R(s_T,a_T) +   V^{LA}(s_T)] \ldots ]. \nonumber
\end{eqnarray}
with discount factor $\gamma=1$. That is, estimated value may make only a relatively small contribution to the value estimation and that estimated value is about future states closer to the end of the game when they exhibit smaller estimation errors.

In the look-ahead model, branching factor was reduced by the following heuristics:
\begin{itemize}
  \item We used a deterministic model and reduced the branching factor since each stochastic event would require a
    branch in the look-ahead tree. Approximations: (i) non-edible ghosts are always chasing Ms.~Pac-Man,
    (ii) edible ghosts are fleeing away from Ms.~Pac-Man, and (iii) Ms.~Pac-Man may stop at positions,
    where it used to perform zig-zagging.
  \item No agent can turn back to a previously visited position. It renders the state graph
    to a simple tree --~where backtracking is not possible~-- reducing branching factor further.
  \item The trajectory is not evaluated further in case of any event that resets the model, such as the death
    of Ms.~Pac-Man agent.
\end{itemize}
Tables~\ref{t:LA} and \ref{t:LA_gain} show the results.

\begin{table}[h!]
    \centering
    \begin{tabular}{ p{25mm} c c c c c c c c c c c c }
        \noalign{\smallskip}\hline\noalign{\smallskip}
            Look-ahead & none & 1 & 2 & 3 & 4 & 5 & 6 &  7 & 8 & 9 & 45  \\
            Average score & 6680 & 6089 & 4702 & 4677 & 4622 & 6013 & 6266 & 6533 & 6561 & 6601 &  \textbf{7130}  \\
        \noalign{\smallskip}\hline
    \end{tabular}\smallskip\smallskip
    \caption{The effect of look-ahead on the score for 3 lives. Single level game, which is over when all dots are eaten.}\label{t:LA}
    \vskip -6pt
\end{table}
\begin{table}[h!]
    \centering
    \begin{tabular}{ p{25mm} c c c }
        \noalign{\smallskip}\hline\noalign{\smallskip}
            Look-ahead & none &  45  \\
            Avg. life lost & 2.82 & 1.25  \\
            Avg. score/life  & 2368 & 5704  \\
        \noalign{\smallskip}\hline
    \end{tabular}\smallskip\smallskip
    \caption{The effect of look-ahead on the average life lost out of 3 lives. Average score per life increased by about a factor of 2.4 using look-ahead. Single level game, which is over when all dots are eaten.}\label{t:LA_gain}
    \vskip -6pt
\end{table}

For this illustration we used a single level game. The optimization algorithm did not make use of the information that a life (out of three) is lost, so Ms.~Pac-Man could make sudden transitions in certain cases. Score per life improved considerably by look-ahead, it increased from about 2,400 (no look-ahead) to 5,700, i.e, by more than a factor of 2. In turn, lookahead helped Ms.~Pac-Man to escape traps. Overall score improved from 6680 to 7130 for a relatively high lookahead value. This is so, because score can be increased considerably if more ghosts are catched after eating a single power dot and deep look-ahead is needed to judge this. There is an interesting interplay between categorical perception and look-ahead, since a decision surface may work well: the only relevant variable (apart from the look-ahead estimated value) is the ghost density, which is a single dimension, so example based learning could work here. The learned decision surface would thus implicitly measure ghost density, which could be used in other situations, too.

Lookahead decreases performance if it is shallow. This is due to the non-Markovian state value estimation that overrides inherited behavior. The estimated value of eating the power dot is high due to the zig-zagging behavior that have emerged. However, lookahead trusts value estimation, overrides zig-zagging, and that decreases the score. Deeper look-ahead overcomes this problem.

\section{Discussion}\label{s:disc}

We summarize the main features of the architecture, relate them to RL algorithms and consider the learning demands. Certain
cognitive aspects will also be mentioned. We analyze the architecture from the point of view of evolutionary, individual,
and social learning.

We emphasize that fFSM is very easy to learn in the Ms.~Pac-Man game and it is very simple to run. The memory and computer
time demand of look-ahead depends on the branching ratio and the size of the state space. fFSM,
however, works in the space of factored variables and may considerably reduce the exponent of the size of the state space of RL.

We have given an example that feature extraction is hard: the machine learning algorithm should figure out high level concepts, like ghost density, to make a correct decision if it should eat the power dot or, instead it should escape in order to attract more ghosts to another power dot. Such extraction of low-dimensional observations, or features is an unsolved problem in machine learning for the general case. Ghost density is an illuminating example: we argued that TD error can be used to learn the decision surface and the decision will reflect ghost density. The parameter `ghost density' could be used in other scenarios, too. However, in the general case it is unclear how to separate this factor from the power dot, or the particular structure of the labyrinth where the decision is to be made.

Modules: modules are spatial-temporal entities. Consider playing volleyball: in this scenario, there are multiple modules and
accompanying activities, such as catching the ball and running, which involves arm and body modules. Ms.~Pac-Man is simple from this point of view, we did not have to consider problems like the decoupling of catching the ball and running. Such problems, however, may pose considerable challenge for other applications. If arm and body modules can be properly decoupled, then look-ahead can take advantage of the factored FSM description in both domains. In the example above, the FSM may be written as a rule: ``\texttt{if Run+ and CatchTheBall+ then \texttt{PassTheBall+}}''. An analog example for Ms.~Pac-Man is ``\texttt{if DotOn+ and IStepOnIt+ then \texttt{DotOn-}}''.

In order to illustrate the problems of missing information we showed two examples in Fig.~\ref{f:pacmantrap}. In one situation (Fig.~\ref{f:Trap1}), Ms.~Pac-Man is waiting on the wrong side of the power dot. In the other case (Fig.~\ref{f:Trap2}), Ms.~Pac-Man is chasing and edible ghost and is surrounded by non-edible ones. For the human observer, these are simple cases. But the situation becomes different also for human observers if only the odor of the power dot and the odor of the ghosts are available. We note that for such poor sensory systems, temporal depth could hardly help, whereas TD error would clearly indicate that there are occasions that largely differ. Additional (e.g., visual) information is needed to build the fFSM in order to use look-ahead.

The fFSM based description has a number of advantages: fFSM model can be the zeroth approximation and can be enriched, e.g.,
stochastic aspects can be included, if more information becomes available. Also, fFSM model offers considerable improvements at moderate computer time requirements. For example, changes in the labyrinth can be incorporated, new FSMs like doors, which can be open or closed can be included.

An fFSM becomes costly if branching ratio is large. In this case, sampling methods can be used, but look-ahead becomes time-consuming making look-ahead useless under time constraints. Furthermore, look-ahead will not work in partially observed worlds when value estimation may be very poor. Cognitive aspects include certain forms of autism. In autism, sensing of the `partner's mind' is sometimes spoiled; emotion detection or emotion understanding are weak. Then, the lack of information about the emotional state and the intentions of the partner makes computational costs high since branching ratio becomes high. At the same time, estimation of rewards is spoiled under such circumstances.

The fFSM description is advantageous in event-learning formalism described in \cite{Szita03Epsilon-MDPs}. In event-learning optimization decides about the desired next state creating a subproblem in the RL task. This subproblem is given to a backing controller, which tries to solve it. Beyond the advantage that desires may be factored, event-learning supports flexible definition of time steps, can utilize backing robust controllers making use of crudely learned inverse dynamics and thus it fits the module description we used here.

State description as the list of available modules and choice of action as the choice of a module was first described in \cite{Kalmar98Module-Based}. The choice of a single module was extended to the choice of low-complexity modules, the macros in \cite{szita07learning} by means of slow, evolution-like, globally optimizing policy search, the CEM.

Low-complexity macros might serve RL since they are factored. If states defined by such macros is Markovian then large RL
problems can be optimized since learning time will scale polynomially if sampling is used \cite{szita08factored}. A further advantage appears if the model, i.e., the transition probability matrix is learned since then the exploration--exploitation dilemma is resolved \cite{szita09optimistic-short}.

With regards to macros, we have found experimentally that (i) the number of optimized macro states can be very small, but (ii) these states may be non-Markovian. Non-Markovity can be early detected by measuring the distribution of the TD errors in each state, which is easy. One may try to optimize each temporally extended macro states by applying policy optimization during the time the macro is used followed by greedification.

Extension of state description by temporal depth is not feasible in general, since temporal depth
comes to the exponent of the size of the state space. For example, in Ms.~Pac-Man huge temporal
depths are required to improve cumulated rewards (Table~\ref{t:LA}). On the other hand, fFSM approximations may be easy to learn, look-ahead
comes cheap for fFSM models and even crude approximations may improve performance. Look-ahead may
also improve value estimations, since it works by computing immediate rewards and then adds
the estimated (and discounted) values of the inherited macro states.

From the point of view of cognition, inherited CEM optimized policy provides inherited values for the behavioral states. Look-ahead can decrease the error of these inherited values in the actual world since it estimates the rewards and discounts the effect of the inherited policy. Look-ahead thus changes behavior. However, inherited rules and related behaviors may reappear  by aging if working memory degrades since look-ahead relies on working memory.

Further work is needed to provide a solid mathematical framework for the \emph{look-ahead extended approximately deterministic fFSM description} (LE-fFSM). LE-fFSM is promising since look-ahead can diminish non-Markovity and then factored RL methods with attractive scaling properties may apply. However, the feature extraction problem that suits LE-fFSM is largely unsolved. Neurally motivated hierarchical exact matrix completion on spatio-temporal representations \cite{lorincz11Markov} seem appealing in this respect.

\subsection{Evolutionary, individual and social learning}

It has been a long-standing problem how to distinguish evolutionary, individual and social learning
\cite{gilbert06emerging}. Our architecture offers the following nomenclature:
\begin{description}
  \item[Evolutionary learning] is a selective globally optimizing method that gives rise to
  low-complexity rules, i.e., macros and behavioral states. The result of evolutionary learning is simple
  and is easy to pre-wire.
  \item[Individual learning] learns TD errors, optimizes modules, e.g., fine tunes the thresholds
  of the rules, computes values and TD errors of macro-states, may increase state space by decision
  surfaces, learns factored (stochastic) FSM, applies look-ahead to predict the behavior, learns to
  control the individual fFSMs (e.g., switches them on and off), may learn the inverse dynamics and apply
  robust controller to improve control and to solve subtasks.
  \item[Social learning] requires that the state of other agents including their value estimations be
  sensed, so other agents become fFSMs in the model. The lack of this knowledge, however, means
  that stochastic FSM agents with hidden variables enter the optimization problem. Such tasks are hard, value estimation may become highly imprecise, and may prohibit collaboration and social learning, especially since other agents are not stationary, they learn \cite{lorincz07mind}.
\end{description}

\subsection{The concept of state in MDPs}\label{s:state}

Close-to-deterministic factored finite-state machine description seem attractive for MDPs, since Laplace's demon is a good initial approximation of the world for different reasons \cite{wolpert08physical,song10limits} and fFSMs are tractable from the point of view of problem size in reinforcement learning. However, the extraction of close-to-deterministic variables (features) is an unsolved problem. Low-complexity, temporally extended and close-to-deterministic factors can satisfy the concept of states in MDPs to a large extent. The environment as well as the agent's action may switch the factors ON and OFF and it would correspond to state-action-state transitions. Controlled optimal switching is then the task of MDPs, which is the concept of event-learning \cite{Szita03Epsilon-MDPs}.

\section{Conclusions and outlook}\label{s:concout}

We considered spatio-temporal modules in this paper. We showed how to make the transition from the modular description to state description and how to find out if states are Markovian or not. We treated different options for achieving the Markovian property and suggested that the learning of factored FSM is the key since it suits look-ahead based improvement of value estimation. We also argued that learning of the factors in the Ms.~Pac-Man problem happens to be easy and noted that the general problem is unsolved in machine learning. We introduced look-ahead into the architecture by means of a deterministic and approximate fFSM, and conjectured that factored RL models with fFSM look-ahead have many desirable features, including close to Markovian description, polynomial scaling and diminishing the exploration-exploitation dilemma by model learning.

We sketched some cognitive aspects of the model; one of the origins of autistic behavior and changes of behavior if the capacity of working memory changes.

We considered the architecture from the point of view of evolutionary, individual and social learning and tried to separate key aspects of this long-standing problem. We have shown that CEM may give rise to simple low-complexity policies, which are easy to inherit. A particular property of individual learning is to construct fFSM, use it for look-ahead in order to diminish errors of value estimation and the non-Markovian characteristics of inherited macro states. Social learning, in our framework, depends on sensory information; if values or TD errors of the partners can be sensed then fFSM description may be sufficient and collaborative planning may be eased considerably.

\section{Acknowledgement}
Thanks are due to \'Akos Bontovics and to M\'ark Farkas for their help and support during the course of this work. The European Union and the European Social Fund have provided financial support to the project under the grant agreement no. T\'AMOP 4.2.1./B-09/KMR-2010-0003.

\clearpage
\end{document}